\documentclass[conference]{IEEEtran}
\IEEEoverridecommandlockouts 
\usepackage[letterpaper,
            top=0.95in, bottom=0.75in,
            left=0.7in, right=0.7in,
            headheight=0pt, headsep=0pt]{geometry}
            
\usepackage[T1]{fontenc}
\usepackage[utf8]{inputenc} 
\usepackage{cite}
\usepackage{amsmath,amssymb,amsfonts}
\usepackage{algorithm}
\usepackage[noend]{algpseudocode}
\usepackage{graphicx}
\usepackage{booktabs}
\usepackage{enumitem}
\usepackage[hidelinks]{hyperref}
\makeatletter
\newcommand{\algmargin}{\the\ALG@thistlm}
\makeatother
\newlength{\whilewidth}
\algdef{SE}[parWHILE]{parWhile}{EndparWhile}[1]
  {\parbox[t]{\dimexpr\linewidth-\algmargin}{%
     \hangindent\whilewidth\strut\algorithmicwhile\ #1\ \algorithmicdo\strut}}
  {\algorithmicend\ \algorithmicwhile}

\algdef{SE}[IF]{parIF}{EndparIF}[1]
  {\parbox[t]{\dimexpr\linewidth-\algmargin}{%
     \hangindent\whilewidth\strut\algorithmicif\ #1\ \algorithmicthen\strut}}
  {\algorithmicend\ \algorithmicif}

\algtext*{EndparIF}

\algnewcommand{\parState}[1]{\State%
  \parbox[t]{\dimexpr\linewidth-\algmargin}{\strut #1\strut}}
\def\BibTeX{{\rm B\kern-.05em{\sc i\kern-.025em b}\kern-.08em
    T\kern-.1667em\lower.7ex\hbox{E}\kern-.125emX}}
    
\begin{document}

\title{Gaussian Processes UCB for decentralized coverage control under unknown density}


\author{
\IEEEauthorblockN{Gennaro Guidone\IEEEauthorrefmark{1},\;
Luca Monegaglia\IEEEauthorrefmark{1},\;
Elia Raimondi\IEEEauthorrefmark{1},\;
Han Wang,\;
Mattia Bianchi,\;
Florian Dörfler}

\thanks{\IEEEauthorrefmark{1} Authors contributed equally.

The authors are with the Automatic Control Laboratory (IfA), ETH Zürich, Zürich, Switzerland. \{gguidone, lmonegaglia, eraimondi, hanwang1, mbianch, dorfler\}@ethz.ch
 }}
\maketitle

\begin{abstract}
We present a novel decentralized algorithm for coverage control in unknown spatial environments modeled by Gaussian Processes (GPs). To trade-off between exploration and exploitation, each agent autonomously determines its trajectory by minimizing a local cost function. Inspired by the GP-UCB (Upper Confidence Bound for GPs) acquisition function, the proposed cost combines the expected locational cost with a variance-based exploration term, guiding agents toward regions that are both high in predicted density and model uncertainty. 
Compared to previous work, our algorithm operates in a fully decentralized fashion, relying only on local observations and communication with neighboring agents. In particular, agents periodically update their inducing points using a greedy selection strategy, enabling scalable online GP updates. We demonstrate the effectiveness of our algorithm in 
simulation.
\end{abstract}

\section{Introduction}
Multi-robot systems are gaining increasing prominence due to their ability to collaboratively execute tasks that are beyond the capabilities of individual robots. In this context, we focus on the problem of \emph{Multi-Agent Coverage Control}, where a team of communicating robots coordinates their positions to optimally cover a given spatial domain. This coverage is guided by a scalar-valued \emph{density function}, which encodes the relative importance of different regions in the environment, with higher density indicating higher priority for sensing or monitoring. A landmark contribution in this field is \cite{cortes2002}, which addressed the coverage problem under the assumption that the density function is globally known. In this setting, each agent improves coverage performance by navigating toward the centroid (mass center) of its assigned Voronoi cell, computed with respect to the known density.

However, this fully informed scenario is unrealistic when the density function is unknown a priori. Therefore, subsequent research has focused on scenarios in which the agents must estimate the density function from local measurements, while simultaneously optimizing their positions for effective coverage\cite{Schwager2007}, \cite{He2023}. These approaches require balancing exploration, acquiring sufficient information about the environment, and exploitation, moving to maximize a coverage function.

 Early approaches to estimating the density function relied on parametric models, such as linear combinations of basis functions \cite{schwager2008,schwager2017}. More recently, \emph{Gaussian Processes} (GPs) have emerged as a powerful and flexible nonparametric alternative for modeling spatial fields in coverage applications \cite{Viseras2016,Krause2008NearOptimal,Luo2018}. In parallel, several works have focused on providing theoretical guarantees for multi-robot GP coverage \cite{Wei2021}. Recently, sparse and variational GP frameworks have been adopted to enable decentralized coverage with scalability improvements \cite{Cao2025}. 

While GPs provide high-fidelity estimates and quantify uncertainty, most implementations in the coverage control literature remain centralized \cite{benevento2020, carron2015}. Centralized methods benefit from a global view of the explored environment and agent states. On the downside, they are computationally expensive and become impractical for large-scale deployments. Only a limited number of works consider decentralized GP-based solutions \cite{Luo2018,nakamura2022}. However, these typically rely on each agent sharing all observations with its neighbors. As a result, the effective model complexity per agent remains equivalent to the centralized case, undermining scalability benefits.

In this paper, we propose a GP-based distributed algorithm for
coverage control under uncertainty. Each agent keeps its own GP to approximate the density locally. To balance exploration and exploitation, the agents move based on local cost functions that pair a GP-UCB–style acquisition term \cite{srinivas2012} with the classical locational coverage cost. To improve scalability, we use sparse GPs, with inducing points chosen based on the information gain via a greedy strategy.  
Furthermore, to keep estimates aligned across agents, we let the agents cooperate by sharing information over a communication graph, and averaging the estimated kernel hyperparameters using a consensus protocol. The advantage is that the agents do not have to share all their observations with their neighbors, resulting in a substantial complexity reduction.  
We show in simulation that our algorithm outperforms the centralized GP-based state-of-the-art method in \cite{benevento2020}. Our results also show that the presence of exploration can allow the agents to escape local minima, ultimately improving coverage efficiency even compared to model-based methods such as \cite{cortes2002}. 

\section{Problem Formulation and Preliminaries} \label{sec:preliminaries}

We start by reviewing the coverage control problem and notations from \cite{cortes2002}, and some GPs preliminaries. 

\subsection{Coverage Control Problem Setting}\label{sec:CoverageControl}

Let $n$ robots operate in a compact, convex planar domain $Q \subset \mathbb{R}^2$. Denote the position of robot $i$ at time $t$ by $p_i(t)$, and let $\mathcal{P} = (p_1, \dots, p_n) \in Q^n$ be the configuration of the robot team. Each robot follows first-order dynamics $\dot{p}_i = u_i$, where $u_i$ is the control input, constrained as
\begin{equation}
    u_i \in U := \{ u \in \mathbb{R}^2 \mid \|u\| \leq v_{\text{max}} \}.
\end{equation}

The domain $Q$ is partitioned using a Voronoi diagram based on robot positions, defined as $\mathcal{V}(\mathcal{P}) = \{V_1(\mathcal{P}), \dots, V_n(\mathcal{P})\}$, 
\begin{equation}
    V_i(\mathcal{P}) = \left\{ q \in Q \mid \|q - p_i\| \leq \|q - p_j\|, \quad \forall j \neq i \right\}.
    \label{voronoiPartition}
\end{equation}
Each robot is responsible for sensing within its cell $V_i(\mathcal{P})$.
The communication topology is described by the Delaunay graph \cite{cortes2004}, where edges connect robots whose Voronoi cells share a boundary. We define the  neighbor set of robot $i$, 
\begin{equation}
    \mathcal{N}_i := \left\{ j \mid V_i \cap V_j \neq \emptyset \right\}.
\end{equation}
 the adjacency matrix $A \in \mathbb{R}^{n \times n}$, with  
\begin{equation}
    A_{ij} = 
    \begin{cases}
        1, & \text{if } j \in \mathcal{N}_i, \\
        0, & \text{otherwise},
    \end{cases}
\end{equation}
and    the diagonal \emph{degree matrix} $D\in\mathbb R^{n\times n}$ with entries $D_{ii} = \sum_{j=1}^n A_{ij}$. Then, the \emph{graph Laplacian} is defined as:
\begin{equation}
    L = D - A.
    \label{eq: Laplacian}
\end{equation}
Note that the Voronoi diagram, communication graph, Laplacian matrix depend on the position of the robots, hence they are time-varying. 

The relative importance of each location is modeled by an unknown density function $\phi: Q \rightarrow \mathbb{R}_{+}$. Higher values of $\phi(q)$ indicate regions that are more critical to monitor, e.g., $\phi$ could encode the spatial distribution of wildfire risk in a forest or the spatial distribution of human presence in a search-and-rescue scenario.

As in \cite{cortes2002}, we assume that the quality of sensing degrades with distance; we describe this degradation with a non-decreasing differentiable function $ f: \mathbb{R}_+ \rightarrow \mathbb{R}_+$. 
Assuming that each robot is responsible for sensing within its Voronoi region, the total sensing cost over the entire environment is captured by the following locational cost function:
\begin{equation}\label{eq:cost}
    \mathcal{H}(\mathcal{P}, \mathcal{V(P)}) = \sum_{i=1}^n \int_{V_i(P)} f(\|q - p_i\|) \phi(q) \, dq.
\end{equation}
For instance, fixing the choice of $f$ to $f(\|q-p_i\|) = \frac{1}{2} \|q - p_i\|^2$
results in the locational cost function 
\begin{equation}
    \mathcal{H}(\mathcal{P}, \mathcal{V(P)}) = \sum_{i=1}^n \int_{V_i(P)} \frac{1}{2} \|q - p_i\|^2 \phi(q) \, dq.
\end{equation}
We wish to minimize $\mathcal{H(P)}$ by controlling the robot configuration. The gradient of $\mathcal{H}$ with respect to the robot position $p_i$ is given by (see \cite{cortes2002}):
\begin{equation}
    \frac{\partial \mathcal{H}}{\partial p_i} = -M_{V_i} (C_{V_i} - p_i),
\end{equation}
where
\begin{align}
    M_{V_i} = \int_{V_i} \phi(q) \, dq  \qquad
    C_{V_i} = \frac{1}{M_{V_i}} \int_{V_i} q \phi(q) \, dq
\end{align}
are the mass and centroid of cell $V_i$ under the density distribution $\phi$.

\subsection{Function Estimation via Gaussian Processes}\label{sec:GP}

To estimate the unknown density function $\phi: Q \rightarrow \mathbb{R}_{+}$, we use a Gaussian Process (GP)\cite{Rasmussen2006Gaussian}.

\subsubsection{Gaussian Process Model}

A Gaussian Process is a collection of random variables, any finite subset of which follows a joint Gaussian distribution. Formally, we write:
\begin{equation}
    \phi \sim \mathcal{GP}(\mu, k),
\end{equation}
where $\mu$ is the mean function, and $k$ is the kernel function encoding spatial correlation between two points, see \cite{Rasmussen2006Gaussian}. 

Given a set of measurements $\mathcal{D} = \{(q_i, y_i)\}_{i=1}^N$, where $y_i = \phi(q_i) + \epsilon$ and $\epsilon \sim \mathcal{N}(0, \sigma^2)$, the GP posterior conditioned on $\mathcal{D}$ is
again a GP,
\begin{align}
    \phi \mid \mathcal{D} \sim \mathcal{GP}(\mu_{\mathcal D}, k_{\mathcal D}) 
\end{align}
where
\begin{align*}
    \mu_{\mathcal D}(q) &= \mu(q) + k(q,\mathcal{Q})^\top  (K_{\mathcal{Q}\mathcal{Q}} + \sigma^2 I)^{-1} (Y - \mu(\mathcal{Q})), \hspace{-0.5em} \\
    k_{\mathcal D}(q,q') &= k(q,q') - k (q,\mathcal{Q})^\top (K_{\mathcal{Q}\mathcal{Q}}+ \sigma^2 I)^{-1} k(q',\mathcal{Q}),
\end{align*}
 $K_{\mathcal{Q}\mathcal{Q}} \in \mathbb R^{N \times N} $ is the kernel matrix with entries $K_{ij} = k(q_i, q_j)$, and $k(q,\mathcal{Q}) = [k(q, q_1), \dots, k(q, q_N)]^\top$. For instance, the posterior at a test point $q$ is a Gaussian variable with mean and variance
\begin{align*}
    \mathbb{E}[\phi(q)\mid \mathcal D] = \mu_{\mathcal D} (q), \ \textnormal{Var}[\phi(q)\mid \mathcal D]=  k_{\mathcal{D}}(q,q). 
\end{align*}

\subsubsection{Scalable Inference with Inducing Points}

To make inference tractable for large datasets, we adopt a sparse approximation of the full GP using a set of $M \ll N$ representative \textit{inducing points} $\mathcal{Z} = \{(q_j, y_j)\}_{j=1}^M \subset \mathcal{D}$ (the choice of the inducing points is discussed later on); the inducing points act as a low-dimensional representation of the latent function.  This  method reduces the complexity of training from $\mathcal{O}(N^3)$ in the full GP case to $\mathcal{O}(NM^2)$ \cite{Rasmussen2006Gaussian}, where $N$ is the number of observations and $M$ is the number of inducing points. Instead of conditioning directly on all $N$ observations, we approximate the posterior of the GP by conditioning on $\mathcal{Z}$:
\begin{align}
    \phi \mid \mathcal{Z} \sim \mathcal{GP}(\mu_\mathcal{Z}, k_\mathcal{Z}) 
\end{align}
where $\mu_\mathcal{Z}$ and $k_\mathcal{Z}$ are the posterior of the mean and variance conditioned only on the inducing points.
This method is particularly well suited for online learning in distributed robotic systems, where communication and computation resources are limited \cite{Norton2023}. 


\subsubsection{Sampling Strategy via GP-UCB}

To efficiently balance \emph{exploitation} and \emph{exploration}, often the \emph{Gaussian Process Upper Confidence Bound} (GP-UCB) strategy \cite{srinivas2012} is adopted. This method selects the next query point by maximizing an acquisition function that combines the GP posterior mean and standard deviation:
\begin{equation}\label{eq:UCBacquisition}
q_* = \arg\max_{q \in \mathcal{S}} \left( \mu_\mathcal{D}(q) + \beta^{1/2} \cdot \sqrt{k_\mathcal{D}(q,q)} \right),
\end{equation}
where $\mathcal{S}$ is the set of the selectable points and $\beta > 0$ modulates the trade-off between exploration and exploitation. An appealing property of GP-UCB is its strong theoretical guarantees: under mild assumptions, it has been proven to achieve \emph{sublinear cumulative regret} with respect to the optimal sampling policy \cite{srinivas2012}.

\subsection{Control goal}

In this paper, we consider a multi-agent coverage control problem, modeled as in Section~\ref{sec:CoverageControl}. The density function is unknown, and modeled as a GP as in Section~\ref{sec:GP}. Specifically, the objective is to develop a scalable, decentralized control and estimation algorithm that enables the robots to:
\begin{enumerate}[topsep=0pt, itemsep=0pt, parsep=0pt]
    \item explore the environment to construct an accurate estimate of the unknown density function $\phi$;
    \item arrange themselves to minimize the  sensing cost $\mathcal{H}$.
\end{enumerate}

\section{Methodology}
\label{sec:method}

In this section, we present our proposed method, summarized in Algorithm~\ref{alg:gp-coverage}. This algorithm is fully decentralized, as 
each robot determines its trajectory based solely on its own observations and the information received from neighboring agents.
In particular, 
each robot holds its own Gaussian Process to estimate the density function $\phi$.  The GP prior parameters and inducing points kept by each agent are updated via observations and communications. Then, a local cost function guides the  movement of each robot: independent minimization of these costs, performed  through gradient descent,  determines the agents’ motion dynamics. We start by formalizing the newly proposed cost function, which is the core and main novelty of our algorithm. 

\subsection{Cost function}\label{Loss function}
Inspired by the GP-UCB strategy \eqref{eq:UCBacquisition} and the locational cost $\mathcal H$ in \eqref{eq:cost}, we propose the following cost for agent \( i \):
\begin{equation}
    \mathcal{H}_i'(V_i) = \mathbb{E}[\mathcal{H}_{i}^{GP}(V_i)] + \beta^{1/2} \cdot \sqrt{\mathrm{Var}[\mathcal{H}_{i}^{GP}(V_i)]},
    \label{eq: Hprime}
\end{equation}
where $\mathcal{H}_{i}^{GP}$ is the locational cost  calculated using the GP posterior estimate of agent \( i \): $\phi_i \mid \mathcal{Z}_i \sim \mathcal{GP}(\mu_{\mathcal{Z}_i}, k_{\mathcal{Z}_i})$,  $\mathcal{Z}_i$ is the set of inducing points for agent $i$, 
\begin{align}
\mathcal{H}_{i}^{GP}(V_i) &= \int_{V_i} f_i(q)\phi_i(q) dq 
\\
\mathbb{E}[\mathcal{H}_{i}^{GP}(V_i)] &= \int_{V_i} f_i(q)\mu_{\mathcal{Z}_i}(q) \, dq, \\
\mathrm{Var}[\mathcal{H}_{i}^{GP}(V_i)] &= \iint_{V_i \times V_i} f_i(q) f_i(q')  k_{\mathcal{Z}_i}(q, q') \, dq \, dq', \label{eq: HiGPvar}
\end{align}
and \( f_i(q) := f(\|q - p_i\|) \) for brevity of notation.

The variance term in \eqref{eq: HiGPvar} should encourage exploration. In particular, the variance of the global cost \(\mathcal{H}^{GP}(\mathcal{P})\) would be  given by:
\begin{align}
\mathrm{Var}[\mathcal{H}^{GP}(\mathcal{P})] 
&= \mathrm{Var}\left[\sum_i \int_{V_i} f_i(q)\mu_\mathcal{D}(q) \, dq \right] \nonumber \\
&= \sum_{i,j} \iint_{V_i \times V_j} f_i(q) f_j(q') k_\mathcal{D}(q, q') \, dq \, dq',
\label{eq: HGP complete var}
\end{align}
where $\mathcal{D}$ is  the set of all collected samples in a centralized GP setting. However, incorporating the ``complete'' variance term in \eqref{eq: HGP complete var} into the GP-UCB objective would not result in a decentralized algorithm, as it would necessitate each agent to access global information from all other agents. To maintain decentralization, we restrict the variance term to be agent-specific, as in \eqref{eq: HiGPvar}. 

We emphasize that, unlike GP-UCB which is formulated for reward maximization, our objective \(\mathcal{H}_{i}'(V_i)\) is a \emph{cost function} and is therefore minimized. The desired agent position at each time would thus be given by:
\begin{equation}
    p_i \in \arg \min_{p_i \in V_i} \mathcal{H}_{i}'(V_i).
\end{equation}
It is also worth noting that the Voronoi  configuration \eqref{voronoiPartition} of the environment is directly influenced by the position of each agent $\{p_i\}_{i=1}^N$, and thus the term $\mathcal{H}_{i}'(V_i)$ \eqref{eq: Hprime} depends on the position of each agent. However, in the minimization, we consider only the position of $p_i$ keeping the Voronoi configuration computed with the current agents' positions fixed, as in \cite{cortes2002}. 

\subsection{Intuition and Methodological Insights}\label{sec:spatialinterpretation}

\subsubsection{Spatial Interpretation of the Cost}

The cost in \eqref{eq: Hprime} can be intuitively understood as a spatially informed analog of GP-UCB in \eqref{eq:UCBacquisition}, with one term in the cost that favors exploration and one exploitation. 
The term
\begin{equation}
\int_{V_i} f(\|q-p_i\|)\mu_{\mathcal{Z}_i}(q)\,dq
\label{eq:mean_cost}
\end{equation}
 corresponds to the expected locational cost as introduced in~\cite{cortes2002}, and can be intuitively interpreted as favoring regions with higher predicted density.

Similarly, the uncertainty component 
\begin{equation}
\sqrt{\iint_{V_i \times V_i} f_i(q)\,f_i(q')\,k_{\mathcal{Z}_i}(q,q')\,dq\,dq'}
\label{eq:var_cost}
\end{equation}
 encourages motion toward areas of higher predictive variance. One difference with respect to the GP-UCB strategy is that the use of an integral implies that agents are driven toward \emph{regions} with high uncertainty, rather than isolated points, as discussed below.
 

\subsubsection{Gradient descent vs discrete maximization:}

An important aspect of our algorithm is the use of \emph{gradient descent} instead of discrete maximization. While traditional GP-UCB methodologies \cite{srinivas2012} typically select the next action by maximizing the acquisition function over a set of points $\mathcal{S}$, our approach updates the next point by following the gradient of $\mathcal{H}_{i}'(V_i) $ in \eqref{eq: Hprime}. This makes the algorithm more suitable for our setting,  where the robots sample and move along continuous trajectories in a dense environment $Q$.

In classical GP-UCB, the exploration term is the posterior variance at the query point, which is equal to the observation noise variance at any previously sampled location and, for typical kernels, remains nearly flat in its immediate neighborhood. By contrast, in our formulation, the variance term is defined as the posterior variance at each point weighted by its distance from the centroid, and integrated over the entire region $V_i$, rather than being determined solely by the variance at the agent’s exact location. Consequently, this term does not vanish at the agent’s current position nor in its immediate neighborhood. This integrated formulation ensures a nonzero gradient that consistently directs the search toward areas of higher variance. As a result, gradient descent can effectively steer the agent toward informative regions without resorting to a discrete search, while still maintaining the exploration-exploitation balance fundamental to GP-UCB.

\newcommand{\StateCont}[1][1]{\Statex\hspace*{#1\algorithmicindent}}

\subsection{Algorithm description}\label{sec:algo}

The pseudocode for the proposed method is presented in Algorithm~\ref{alg:gp-coverage}. Next, we provide additional context to clarify its key components. Let us emphasize again that
each agent maintains a local GP model, which is updated over time based on local observations and communication with neighboring agents.

\begin{algorithm}[t]
\caption{Decentralized Coverage Control with GP-based Exploration}
\label{alg:gp-coverage}
\begin{algorithmic}[1]
\State \textbf{Input:} Initial position $p_i(0)$, initial GP prior, initialize $\mathcal{Z}_i \gets \emptyset$, $\mathcal{B}_i \gets \emptyset$, step size $\eta$, update frequency $T$
\For{each time step $t = 0, 1, 2, \dots$}
    \For{agent $i = 1, \dots, N$ \textbf{in parallel}}
        \State Compute Voronoi partitions $\mathcal{V}(\mathcal{P})$
        \parState{Construct local communication graph and Laplacian matrix $L$ from neighboring Voronoi cells}
        \parState{Update GP hyperparameters via consensus with neighbors using \eqref{eq:consensusupdate}}
        \State Observe measurement $y_i = \phi(p_i(t)) + \epsilon$
        \State Append $(p_i(t), y_i)$ to temporary buffer $\mathcal{B}_i$
        \If{$t \bmod T = 0$}
            \parState{Exchange and merge inducing points:
            $\mathcal{Z}_i \gets \mathcal{Z}_i \cup \mathcal{B}_i \cup \bigcup_{j \in \mathcal{N}_i} \mathcal{Z}_j$}
            \parState{Select new local elements with Algorithm~\ref{alg:greedyselection}:
            $\mathcal{Z}_i \gets \mathrm{GreedySelection}(\mathcal{Z}_i)$}
            \parState{Update local GP posterior using current inducing points $\mathcal{Z}_i$}
            \State Clear temporary buffer: $\mathcal{B}_i \gets \emptyset$
        \EndIf
        \parState{Compute descent direction $\nabla \mathcal{H}_i'(V_i)$ (see Appendix~\ref{sec:derivative})}
        \If{plateau is detected, see \eqref{plateau}}
            \State $p_i(t+1) \gets \Pi_Q\!\left[p_i(t) - \eta \, \nabla \mathcal{H}_i'(V_i)\right]$
        \Else
            \State Update $p_i(t+1)$ with Adam optimizer
        \EndIf
    \EndFor
\EndFor
\end{algorithmic}
\end{algorithm}

\subsubsection{Hyperparameter averaging}\label{hyperparams_averaging}To maintain coherence across agents’ models, in \textbf{step~5} a Laplacian-based \eqref{eq: Laplacian} consensus update is performed at every iteration, where we average kernel hyperparameters $\boldsymbol{\theta}_i^{(t)}$ across neighbors to ensure consistent posterior inference in the decentralized setting:
\begin{equation}\label{eq:consensusupdate}
    \boldsymbol{\theta}_i^{(t+1)} 
    = \boldsymbol{\theta}_i^{(t)} + \alpha \sum_{j\in\mathcal{N}_i} a_{ij}\big(\boldsymbol{\theta}_j^{(t)}-\boldsymbol{\theta}_i^{(t)}\big)
\end{equation}
for a stepsize $\alpha>0$. We assume no communication constraints and use the bidirectional Delaunay graph induced by the agents’ positions as the communication topology, which is connected for any finite point set. 

\subsubsection{Gaussian Process update}\label{GP_update}
After collecting an observation, each agent appends the new data point to its local buffer $\mathcal{B}_i$. Every $T$ iterations, each agent updates its set of inducing points and corresponding Gaussian Process (GP) model.  The choice of the period $T$ allows to  improve  computational cost  and runtime efficiency, at the cost 
 temporarily retaining high uncertainty in regions an agent
has recently chosen to explore
Specifically, in \textbf{step 9}, each agent aggregates its own inducing points with those of its neighbors through direct communication. After aggregation, each agent selects a subset of $M$ inducing points that best capture the underlying function across its domain. This selection is performed in \textbf{step 10} using a greedy heuristic based on information gain (Sec. \ref{sec:greedyselection}). \\
Once the new inducing set is selected, in \textbf{step 11} the agent updates its GP posterior accordingly.

\begin{figure*}[t]
  \centering
  \begin{minipage}[t]{0.49\textwidth}
    \centering
    \includegraphics[width=\linewidth]{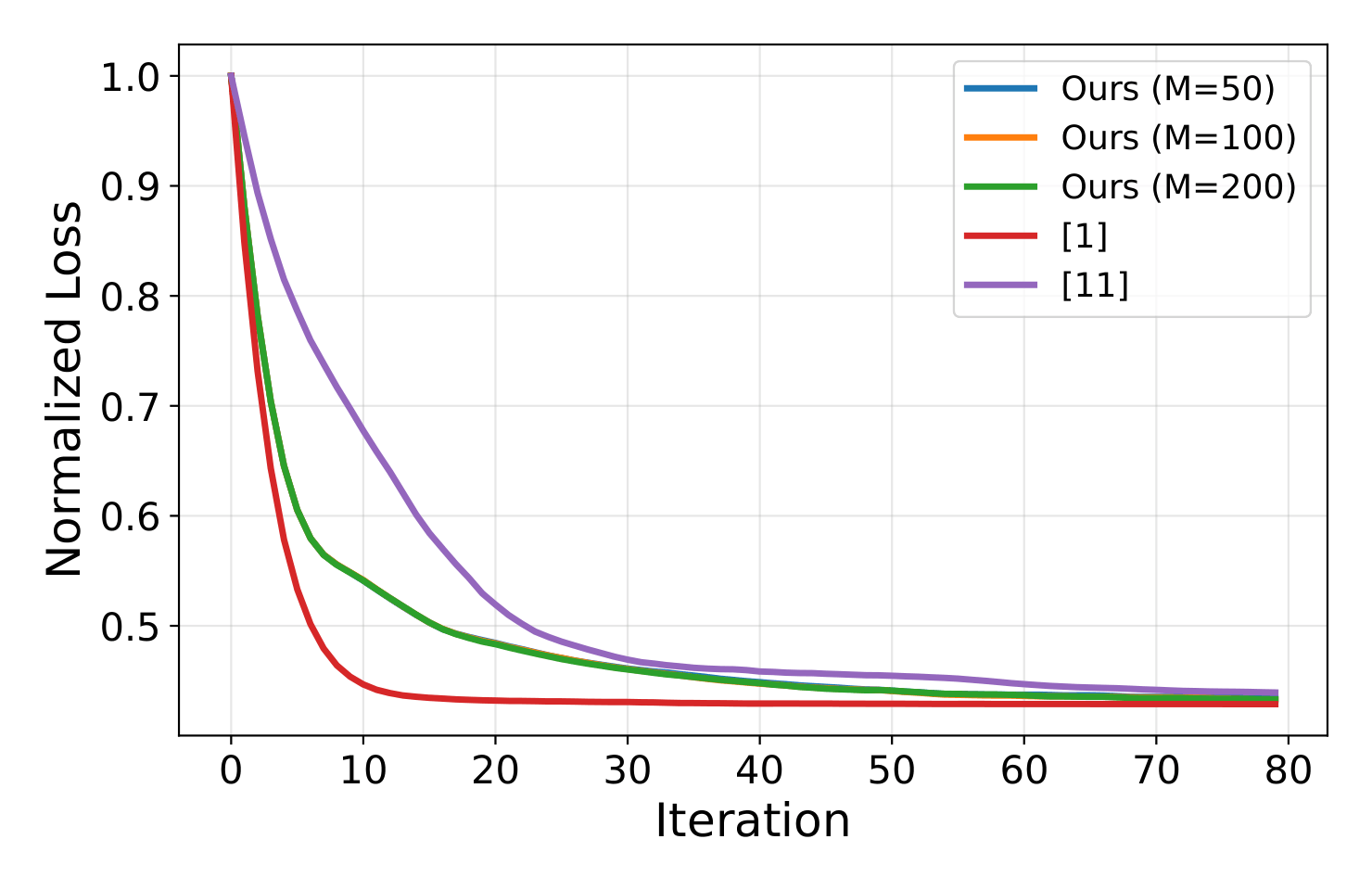}
    \caption{Mean normalized loss}
    \label{fig:mean_loss_benchmark}
  \end{minipage}
  \hfill
  \begin{minipage}[t]{0.49\textwidth}
    \centering
    \includegraphics[width=\linewidth]{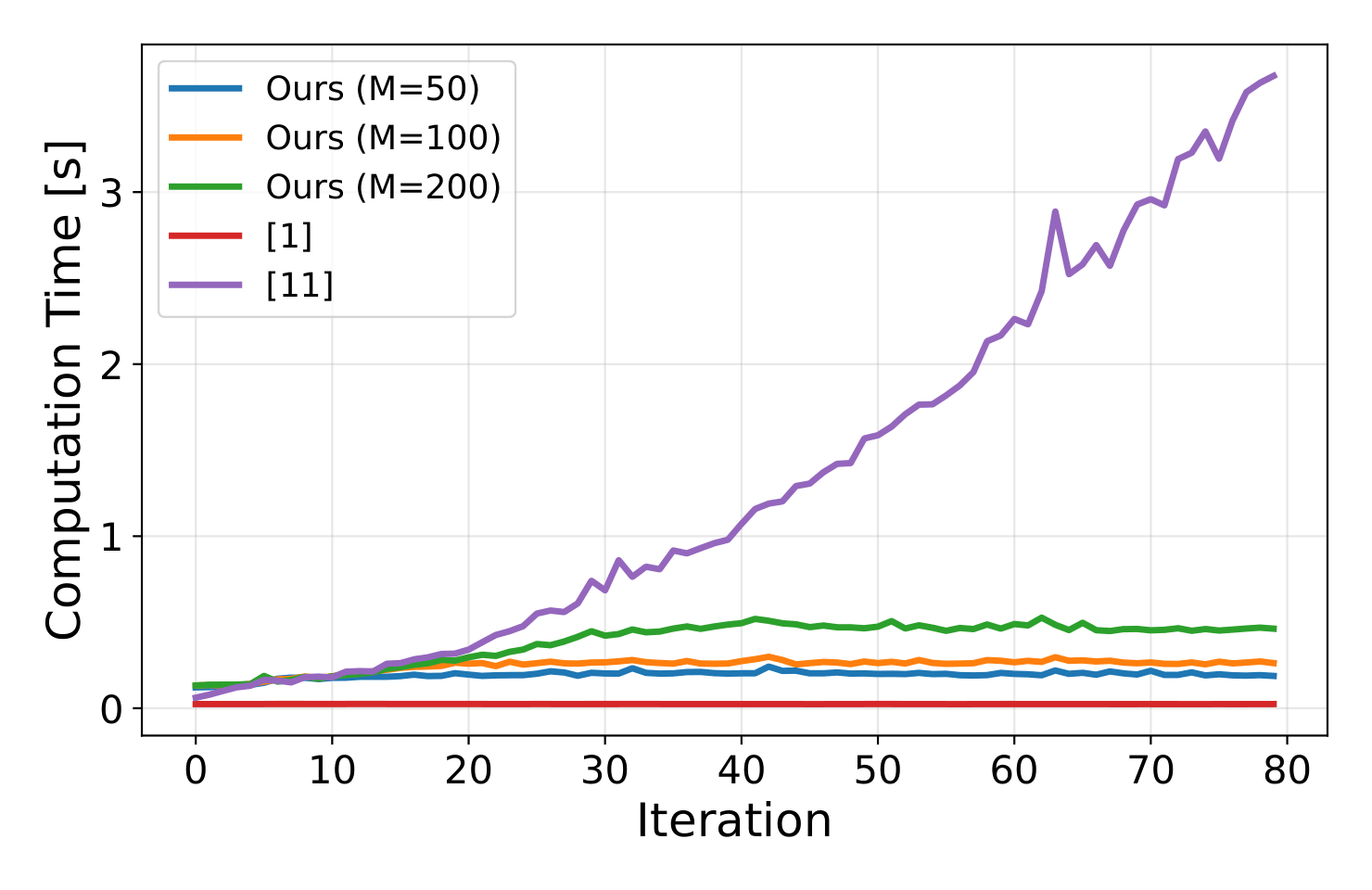}
    \caption{Mean computation time per iteration}
    \label{fig:mean_time_benchmark}
  \end{minipage}
\end{figure*}
\begin{figure}[t]
  \centering
  \includegraphics[width=\columnwidth]{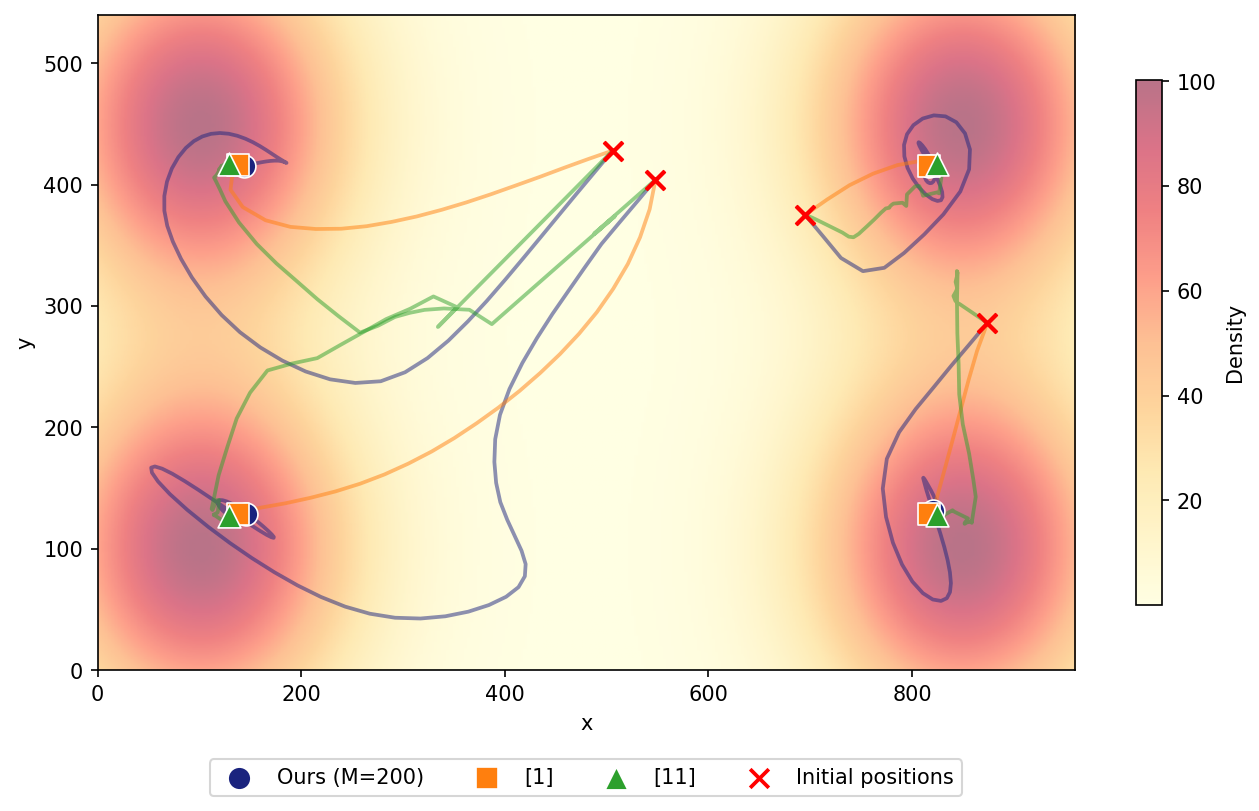}
  \caption{Illustrative comparison of agents' trajectories and final positions in a representative four-agent scenario with four Gaussian modes}
  \label{fig:traj_compare}
\end{figure}

\subsubsection{Gradient descent}\label{Gradient_descent}
The agent moves following the gradient of the loss function \eqref{eq: Hprime} in \textbf{step~16}. For the gradient-based updates, we ultimately employ the Adam optimizer~\cite{kingma2017} in \textbf{step~18},
as it facilitates navigation through flat regions of the objective landscape and accelerates convergence when gradients are small. However, during the initial iterations, the density estimates exhibit considerable fluctuations. In this regime, constant step size and no momentum have been observed to yield a more favorable transient response. \\
The transition to Adam optimization is triggered once a plateau condition is met, defined as the mean relative change in $\sigma_i = \sqrt{\mathrm{Var}[\mathcal{H}_{i}^{GP}(V_i)]}$ over the last $k$ steps remaining below a threshold $\varepsilon$, where $\mathrm{Var}[\mathcal{H}_{i}^{GP}(V_i)]$ is as in \eqref{eq: HiGPvar}.
\begin{equation}
    \frac{1}{k}\sum_{t \in [t' - k, t']} \frac{\|\sigma_{i}(t+1) - \sigma_i(t)\|}{\|\sigma_i(t)\|} \leq \varepsilon 
\label{plateau}
\end{equation}
The computed step is then projected onto the feasible set defined by the spatial constraints, ensuring that the updated position remains within the admissible domain in \textbf{step 16}.

\section{Numerical Results}
\label{sec:experiments}

We evaluate the proposed method on a benchmark composed of 45 episodes arranged as a $3 \times 3$ case matrix with 5 repetitions per case. The nine cases are obtained from the Cartesian product between the number of agents, $n_{\mathrm{agents}} \in \{5,6,7\}$, and three density families: a single centered Gaussian, four corner Gaussians, and a uniform density over the full environment.

The five repetitions associated with each case differ only in the initial agent positions, which are generated from fixed random seeds. We compare three instances of the proposed sparse decentralized method, with inducing-point budgets $M \in \{50,100,200\}$, against the oracle coverage controller of~\cite{cortes2002} and the centralized GP-based method of~\cite{benevento2020}. Performance is measured through the normalized loss, i.e., the loss at iteration $t$ divided by its value at the initial iteration.

Fig.~\ref{fig:mean_loss_benchmark} reports the mean normalized loss aggregated over the full benchmark. The three sparse variants yield almost overlapping loss curves throughout the horizon, indicating that, in the tested range, the coverage performance is insensitive to the inducing-point budget. All three variants outperform the centralized GP baseline~\cite{benevento2020} both during the transient phase and at convergence, while remaining close to the oracle method~\cite{cortes2002}, which achieves the best final metric overall.

Fig.~\ref{fig:mean_time_benchmark} reports the mean computation time per iteration over the full benchmark. For the proposed decentralized method, the reported time is normalized by the number of agents. This choice reflects the underlying computational architecture: each agent performs its own local GP update and control computation on-board, so the overall workload is naturally distributed across the team rather than executed on a single processor. By contrast, the method of~\cite{benevento2020} relies on a single centralized GP model that aggregates all measurements and performs inference centrally; accordingly, its computation time corresponds to a centralized processing pipeline and is not distributed across agents. The proposed method remains computationally bounded over the horizon, with a moderate increase as the inducing-point budget grows, whereas~\cite{benevento2020} exhibits a marked increase in runtime as data accumulate in the centralized GP. As expected, the oracle method~\cite{cortes2002} is the cheapest to evaluate, since it does not require GP inference.

Finally, Fig.~\ref{fig:traj_compare} shows a representative run with four agents in the four-corner Gaussian environment and is meant to qualitatively visualize the motion induced by the different controllers. In particular, it highlights how the agents redistribute themselves across multiple high-density regions and provides an intuitive comparison of the resulting trajectories and final configurations.

\section{Conclusion / Future work} \label{sec:conclusions}

This paper presented a decentralized algorithm that enables a team of agents to simultaneously learn and cover an unknown spatial field by leveraging Gaussian Processes (GPs) with inducing points to reduce computational complexity. A novel loss function was introduced to guide the selection of the next sampling location, explicitly accounting for both exploration and exploitation objectives. Empirical results demonstrate that our method not only surpasses alternative GP-based coverage algorithms in performance but also achieves coverage costs comparable to the algorithm proposed in~\cite{cortes2002}, despite the latter having access to the ground-truth distribution, a significant advantage.
Future work will focus on providing rigorous theoretical guarantees for the proposed method, including establishing asymptotic no-regret bounds. Additionally, incorporating coordination mechanisms that allow agents to anticipate the intended sampling regions of their neighbors, potentially through limited lookahead strategies, could further enhance collective efficiency and coverage performance. Another important direction is to extend the algorithm to the setting where the underlying density function evolves over time, which is highly relevant in dynamic or non-stationary environments.

\bibliographystyle{IEEEtran}
\bibliography{bibliography}

\appendices

\section{Derivation of the Gradient of the Variance Term}
\label{sec:derivative}

Let 
\[
H_i(p_i) \;=\; \int_{V_i} \frac12\,\|q - p_i\|^2 \,\phi(q)\,dq
\]
and denote its variance by
\[
\mathrm{Var}\bigl[H_i(p_i)\bigr]
=\frac14 
\int_{V_i}\!\!\int_{V_i}
\|q-p_i\|^2\,\|q'-p_i\|^2
\;K(q,q')\;
dq\,dq'.
\]
Accordingly,
\[
\sigma_{H_i}(p_i)
=\sqrt{\mathrm{Var}\bigl[H_i(p_i)\bigr]} .
\]

\[
\nabla_{p_i}\sigma_{H_i}
=\frac{1}{2\,\sigma_{H_i}}
\nabla_{p_i}\!\bigl[\mathrm{Var}[H_i]\bigr].
\]

We first compute the gradient of the integrand:
\begin{align*}
\nabla_{p_i}\!&\bigl[\|q-p_i\|^2\,\|q'-p_i\|^2\bigr]
=\\ &-2\,(q - p_i)\,\|q'-p_i\|^2
\;-\;2\,(q'-p_i)\,\|q-p_i\|^2.
\end{align*}
Therefore,
\begin{align}
&\nabla_{p_i} \mathrm{Var}[H_i] \nonumber \\
&=
\frac14
\int_{V_i}\!\!\int_{V_i}
\Bigl[\nabla_{p_i}\!\bigl[\|q-p_i\|^2\,\|q'-p_i\|^2\bigr]\Bigr]
\,K(q,q')\;dq\,dq'
\nonumber\\
&=
-\frac12
\int_{V_i}\!\!\int_{V_i} \Bigl[(q-p_i)\,\|q'-p_i\|^2 \nonumber \\
&\qquad+ (q'-p_i)\,\|q-p_i\|^2\Bigr]
\,K(q,q')\;dq\,dq'.
\end{align}
Finally, inserting into the chain‐rule formula for \(\sigma_{H_i}\) gives
\begin{align}
\nabla_{p_i}\,\sigma_{H_i}
=&
-\frac{1}{4\,\sigma_{H_i}}
\int_{V_i}\!\!\int_{V_i}
\Bigl[(q-p_i)\,\|q'-p_i\|^2 \nonumber \\&+ (q'-p_i)\,\|q-p_i\|^2\Bigr]
\,K(q,q')\;dq\,dq'.
\end{align}

\section{Greedy Selection of Inducing Points}
\label{sec:greedyselection}

Algorithm \ref{alg:greedyselection}, inspired by \cite{Lawrence2002} describes a greedy approach for selecting inducing points in a GP model, tailored for distributed multi-agent coverage tasks. Each agent maintains a set of inducing points, and the goal of the algorithm is to select $m$ inducing points that maximize the information gain, quantified here by the posterior variance at candidate points.

\begin{algorithm}[h]
\caption{Greedy Selection of Inducing Points via Information Gain}
\label{alg:greedyselection}
\begin{algorithmic}[1]
\For{each agent $i$}
    \State Initialize inducing point set $Z_i \gets \emptyset$
    \State Create list $L_i$ containing:
    \Statex \hspace{\algorithmicindent} all inducing points and associated measurements received from Voronoi neighbors
    \Statex \hspace{\algorithmicindent} current inducing points of agent $i$
    \While{$|Z_i| < m$}
        \State Update $K_i^{-1}$ using the Woodbury identity
        \For{each candidate point $u \in L_i \setminus Z_i$}
            \State Compute posterior variance $\sigma_i(u)$ using $K_i^{-1}$
        \EndFor
        \State Select $u^* \gets \arg\max_{u} \sigma_i(u)$
        \State Add $u^*$ to $Z_i$
    \EndWhile
    \State \Return $Z_i$ as the new inducing point set for agent $i$
\EndFor
\end{algorithmic}
\end{algorithm}

\end{document}